\documentclass[10pt,twocolumn,letterpaper]{article}

\usepackage{cvpr}              % To produce the CAMERA-READY version
\usepackage{booktabs}
\usepackage[numbers]{natbib}
\usepackage{soul}
\usepackage{lipsum,microtype}
\usepackage{xcolor}
\usepackage{colortbl}
\usepackage{graphicx}
\usepackage{svg}
\usepackage{amsmath}
\usepackage{amssymb}
\usepackage{booktabs}
\usepackage{multirow}
\usepackage{array}

\definecolor{cvprblue}{rgb}{0.21,0.49,0.74}
\usepackage[pagebackref,breaklinks,colorlinks,citecolor=cvprblue]{hyperref}

\def\paperID{14949} % *** Enter the Paper ID here
\def\confName{CVPR}
\def\confYear{2025}

\title{\textit{TokenMotion}: Decoupled Motion Control via Token Disentanglement for Human-centric Video Generation}

\author{Ruineng Li\textsuperscript{*,1}, Daitao Xing\textsuperscript{*,1}, Huiming Sun\textsuperscript{1}, Yuanzhou Ha, Jinglin Shen\textsuperscript{1}, Chiuman Ho\textsuperscript{1}\\
\textsuperscript{1}OPPO US AI Center\\
{\tt\footnotesize \{ruineng.li2, daitao.xing1, huiming.sun2, jinglin.shen, chiuman\}@oppo.com, hayuanzhou0313@gmail.com}\\
}
\begin{document}

\twocolumn[{
\renewcommand\twocolumn[1][]{#1}
\begin{center}
    \centering
\maketitle
    \includegraphics[width=1.02\textwidth]{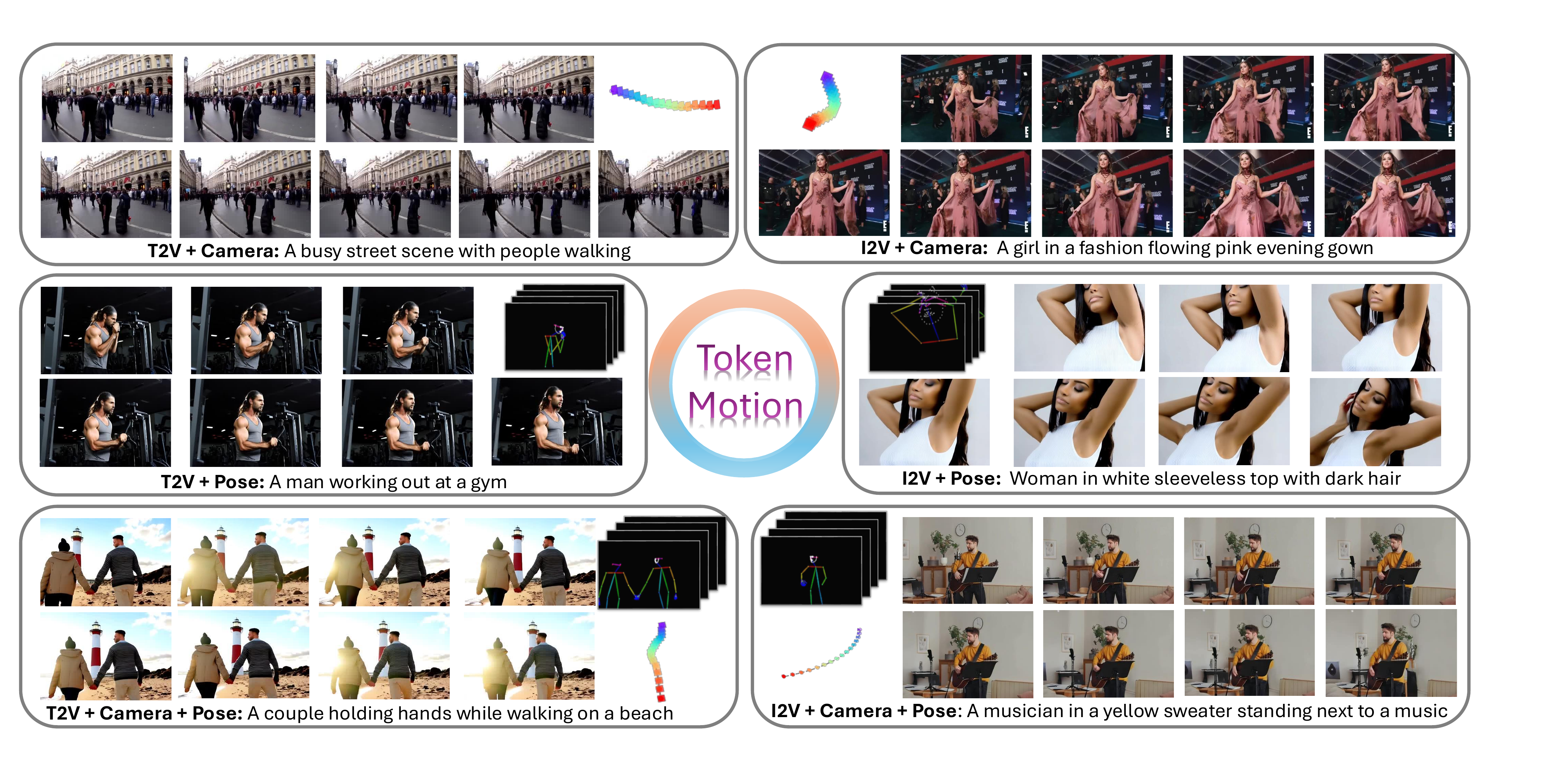}
    \vspace{-10mm}

    \captionof{figure}{\textit{TokenMotion} is a transformer-based video generation framework that enables simultaneous control of camera trajectories and human kinematic patterns. The framework demonstrates versatility across both text-to-video and image-to-video generation paradigms, while supporting flexible control configurations.
    *Text prompts are abbreviated for conciseness.
    }
    \label{fig: Figure1}
\end{center}
}]
\renewcommand{\thefootnote}{\fnsymbol{footnote}}
\footnotetext[1]{Equal contribution.}
% \footnotetext[2]{This work was done when the author was at OPPO US AI Center.}
\begin{abstract}
Human-centric motion control in video generation remains a critical challenge, particularly when jointly controlling camera movements and human poses in scenarios like the iconic Grammy Glambot moment. While recent video diffusion models have made significant progress, existing approaches struggle with limited motion representations and inadequate integration of camera and human motion controls. In this work, we present TokenMotion, the first DiT-based video diffusion framework that enables fine-grained control over camera motion, human motion, and their joint interaction. We represent camera trajectories and human poses as spatio-temporal tokens to enable local control granularity. Our approach introduces a unified modeling framework utilizing a decouple-and-fuse strategy, bridged by a human-aware dynamic mask that effectively handles the spatially-and-temporally varying nature of combined motion signals. Through extensive experiments, we demonstrate TokenMotion's effectiveness across both text-to-video and image-to-video paradigms, consistently outperforming current state-of-the-art methods in human-centric motion control tasks. Our work represents a significant advancement in controllable video generation, with particular relevance for creative production applications.
\end{abstract} 

\section{Introduction}
\label{sec: intro}
% video generation advanecments
% provide more diverse forms of control signals to better improve the controllability of models. 

Video diffusion models have achieved significant progress in text-to-video \cite{VideoCrafter1Chen, LaVie23Wang, ZeroScope23Kha, gao2024lumina} and image-to-video \cite{DynamiCrafter24Xing, SVD23, VideoCrafter24Chen} generation, with recent works incorporating external control signals such as depth maps \cite{SparseCtrl24Guo, wang2024videocomposer}, subject appearance\cite{MotionBooth24Wu, zhao2025motiondirector} and motion \cite{CameraCtrl24He, MotionCtrl24Wang, DragAnything24Wu, DragNUWA23Yin} for better generation controllability. 
Among these, human-centric motion control emerges as a crucial challenge in video generation.
Consider the iconic Grammy Glambot moment, which requires the joint control of both dramatic camera movements and human pose changes - exactly the type of control we aim to achieve. 
The importance of this capability stems from the explosive growth of AI-generated content and its great potential in creative production, from films to viral short-form videos.
% This cruciality stems from the explosive growth of AI-generated content, and its great potentials in creative production from films to viral short-form videos. For instance, it will be highly appealing if users can replicate the iconic Grammy Glambot moment, which requires the joint control of both dramatic camera movements and human pose changes, with their own photographs. 
However, this critical aspect remains largely unexplored, with existing works providing only limited solutions.

% efforts in object motion control and camera control, 
Such limitations in controlling video generation with human-centric motion conditions fall into  two aspects. First, most current works in motion control focus only on either object motion control \cite{Boximator24Wang, DragNUWA23Yin} or camera control \cite{CameraCtrl24He, CamCo24Xu, VD3D24Bah}, lacking the capability of joint control. Second, while few works \cite{MotionCtrl24Wang, DirectAVideo24Yang, MotionBooth24Wu} have attempted to explore jointly controlling camera motion and object motion, they remain limited for human-centric motion control, due to over-simplified motion representations and ad-hoc motion integration: MotionCtrl \cite{MotionCtrl24Wang} uses object-level keypoints, while Direct-A-Video\cite{DirectAVideo24Yang} and MotionBooth\cite{MotionBooth24Wu} uses bounding boxes for object trajectories. These coarse representations fail to capture local movements, especially subtle pose changes. Furthermore, these works all directly integrate two motion signals without handling their potential interactions, and report the presence of motion conflicts in complex joint-control scenarios, indicating this challenge remains unsolved for human-centric motion control. While HumanVid \cite{HumanVid24Wang} shares a similar goal, it focuses on dataset construction and provides only a simple baseline for dataset validation, leaving the technical challenges unaddressed.

A key challenge exists for effectively integrating camera and human motion controls. 
Intuitively, given any pair of video frames, each pixel's motion arises from different combination of camera and human movements. 
This spatial-and-temporally-varying nature of motion composition necessitate unified modeling of both motion signals in the same representation space, along with explicit designs to capture their interactions. 
% To this end, an intuitive approach is to encode every feature in the latent representation with both signals simultaneously and subsequently integrate them. 
% To this end, an intuitive approach is to employ a joint encoding of two types of motion signals on the feature-level, along with dedicated designs for achieving the targeted motion coherence.
However, existing approaches lack such dedicated designs. Moreover, current joint-control methods adopt UNet backbones, which separate object and camera motions into spatial and temporal modeling, respectively, making it suboptimal for joint encoding. Despite DiT-based \cite{Dit23Peebles} alternatives seem a promising choice, joint-control DiT-based models have not yet been proposed. Besides, recent camera control works \cite{VD3D24Bah, Boosting24Cheong} find that previous motion encoding techniques require tailored adaptation for effectively working on DiT-based models, a challenge likely amplified for joint motion control.

% a decouple-and-fuse strategy, to perform human-centric motion control. Concretly, 
In this work, we present \textit{TokenMotion} to address all these challenges, which is the first DiT-based video diffusion model that enables human-centric motion control in video generation.
TokenMotion achieves fine-grained control over all three fundamental scenarios of human-centric motion control: camera-only, human-motion-only, and their joint control. 
At the core of our approach is a novel unified modeling framework that handles camera and human motion through a decouple-and-fuse strategy, bridged by a human-aware dynamic mask for effectively integrating two motion signals. 
Furthermore, to address the challenge of degraded control accuracy in DiT-based frameworks, we introduce motion patchification that encodes motion as fixed-length sequences.
We also demonstrate the generalization capabilities and superiority of our approach across both paradigms of text-to-video and image-to-video, consistently outperforming current state-of-the-art methods. We claim three main contributions summarized as follows:

% TODO: Extensive experiments show that ...
\begin{itemize}
    \item TokenMotion, the first DiT-based video diffusion framework that enables fine-grained control over camera motion, human motion, and their joint interaction, serving a critical problem in AI-generated video content.

    % \item A novel token compression technique that addresses the control accuracy degradation, enabling efficient encoding of motion as fixed-length sequences.
    
    \item A unified modeling framework with a decouple-and-fuse strategy, bridged by a human-aware dynamic mask that effectively handles the spatially-and-temporally-varying nature of combined camera and human motion signals.
    
    \item We demonstrate the generalization capabilities of our approach across both text-to-video and image-to-video paradigms through extensive experiments, consistently outperforming current state-of-the-art methods in human-centric motion control.

\end{itemize}

\section{Related Works}
\label{sec:related work}
\begin{figure*}[t]
  \centering
  \includegraphics[width=0.99\textwidth]{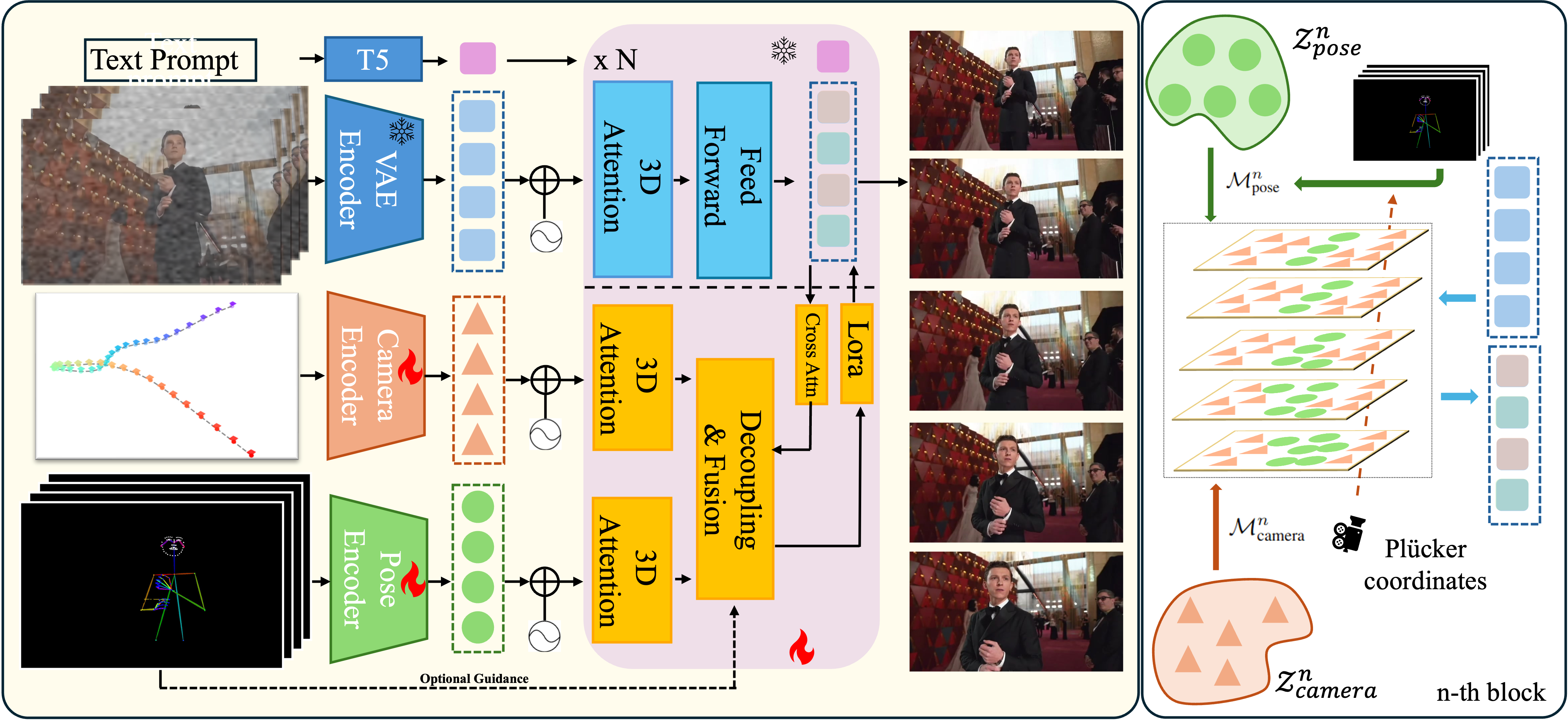}
   \caption{\textbf{Architectural Overview.} TokenMotion presents a novel video generation framework that combines a transformer-based video diffusion model with content-aware motion guidance. The architecture employs dual motion encoders that extract spatio-temporal motion tokens. These motion features are then processed through a specialized decoupling and fusion module, which dynamically modulates the strength of motion guidance based on content characteristics, enabling fine-grained control over temporal consistency.}
   \label{fig:framework}
\end{figure*}

\subsection{Video Diffusion Models}
The domain of video generation \cite{VDM22Ho, Gen1, VideoLDM23, SVD23, LaVie23Wang, ZeroScope23Kha} is undergoing remarkable evolution driven by the advanced generative power of diffusion models \cite{DDPM20Ho, LDM22Rom, Imagen22Sah}. Current approaches primarily utilize UNet-based diffusion backbones to generate videos, including works in text-to-video (T2V) generation \cite{VideoCrafter24Chen, AnimateDiff24Guo, LaVie23Wang, ZeroScope23Kha}, and image-to-video generation (I2V) \cite{I2VGen23Zhang, DynamiCrafter24Xing, SVD23}. To further improve the scalability and the generation capabilities of diffusion models, more recent works are shifting their attention towards transformer-based backbones \cite {Dit23Peebles, WALT23Gupta, sora24Brooks, OpenSoraPlan24Yuan, CogVideoX24Yang}. One of the most recent works, CogVideoX \cite{CogVideoX24Yang} further advances by directly operate in the 3D latent space through the encoding of a 3D Variational Autoencoder (VAE), and a progressive training technique to generate high-quality videos. Besides, another significant line of research focuses on enhancing the generation controllability of video diffusion models through control signals such as dense map \cite{Densemap23Esser, Densemap23Yang}, subject appearance \cite{revideo24Chong, CustomCrafter24Wu, StillMoving24Chefer} and motion \cite{AnimateAnyone24Hu, DragNUWA23Yin, MotionCtrl24Wang, CameraCtrl24He, Boximator24Wang}. 

% \begin{figure*}[t]
%   \centering
%   \includegraphics[width=1.02\textwidth]{figures/Framework.pdf}
%    \caption{\textbf{Overview of architecture.} TokenMotion integrates a transformer based video diffusion model with novel content-aware motion guidance. The training components consists of two motion encoders to extract motion tokens and a decoupling and fusion module for motion distanglment and fusion }
%    \label{fig:framework}
% \end{figure*}

\subsection{Motion Control in Video Generation}
Recent works in motion control introduce different types of motion signals to gain more accurate controllability in video generation, especially for object motion \cite{Boximator24Wang, DirectAVideo24Yang, DragAnything24Wu, DragNUWA23Yin, DragAPart24Li, MagicAnimate24Xu, AnimateAnyone24Hu, Champ24Zhu} and camera motion \cite{CameraCtrl24He, CamCo24Xu, CamI2V24Zheng, VD3D24Bah, Boosting24Cheong}. CameraCtrl \cite{CameraCtrl24He} pioneered in using Pl\"ucker Embedding \cite{sitzmann2021light} instead of start-and-end camera shifts \cite{MotionCtrl24Wang, MotionBooth24Wu}, enabling fine-grained control over intermediate camera poses throughout the trajectories. MagicAnimate \cite{MagicAnimate24Xu} and AnimateAnyone \cite{AnimateAnyone24Hu} utilizes skeletons to enable subtle pose changes beyond object-level motion. However, most works only focus on either type of motions, limiting the scope of generation controllability. Some recent studies take first steps toward jointly controlling object motion and camera motion \cite{MotionBooth24Wu, MotionCtrl24Wang, DirectAVideo24Yang}. MotionCtrl \cite{MotionCtrl24Wang} and Direct-A-Video \cite{DirectAVideo24Yang} adopt a data-driven decoupling approach, which use different data to learn different motion concepts. However, both methods claim to show motion conflicts. Related to these works is ImageConductor\cite{ImageConductor24Li}, which decouples camera motion and object motion for better individual control due to their entanglement in training data, but doesn't explore their joint operation.

Another line of research proposes that the limited motion understanding these current models show originates from the inferiority of their UNet-based backbones \cite{VideoCrafter24Chen, AnimateDiff24Guo, SVD23}, which are claimed to have weaker scalability and limited generative power compared to DiT-based counterparts \cite{Dit23Peebles, sora24Brooks}. Recently, VD3D \cite{VD3D24Bah} introduced the first camera-controlled DiT-based models. Currently, joint-control DiT-based models have not yet been proposed.
\section{Methodology}

% This section presents TokenMotion, a novel approach for generating motion-controlled videos from textual or visual inputs while preserving both camera movements and human motion patterns without distorted artifacts. 
%This section presents TokenMotion, a novel approach that enables fine-grained control over both camera motion and human motion in video generation.
%The overview of the proposed approach is shown in Fig. \ref{fig:framework}. 
%We utilizes two specialized encoders to derive camera patch tokens and human motion tokens independently. 
%Subsequently, the proposed content-aware motion attention module is conducted on the flatten decoupled motion representations to progressively collect camera cues, and human motion cues, generating context-aware motion tokens. The final stage involves updating visual tokens through cross-attention mechanisms applied to the fused motion tokens, followed by adaptation layers to refine the output.

\begin{figure*}
  \centering
  \includegraphics[width=1\textwidth]{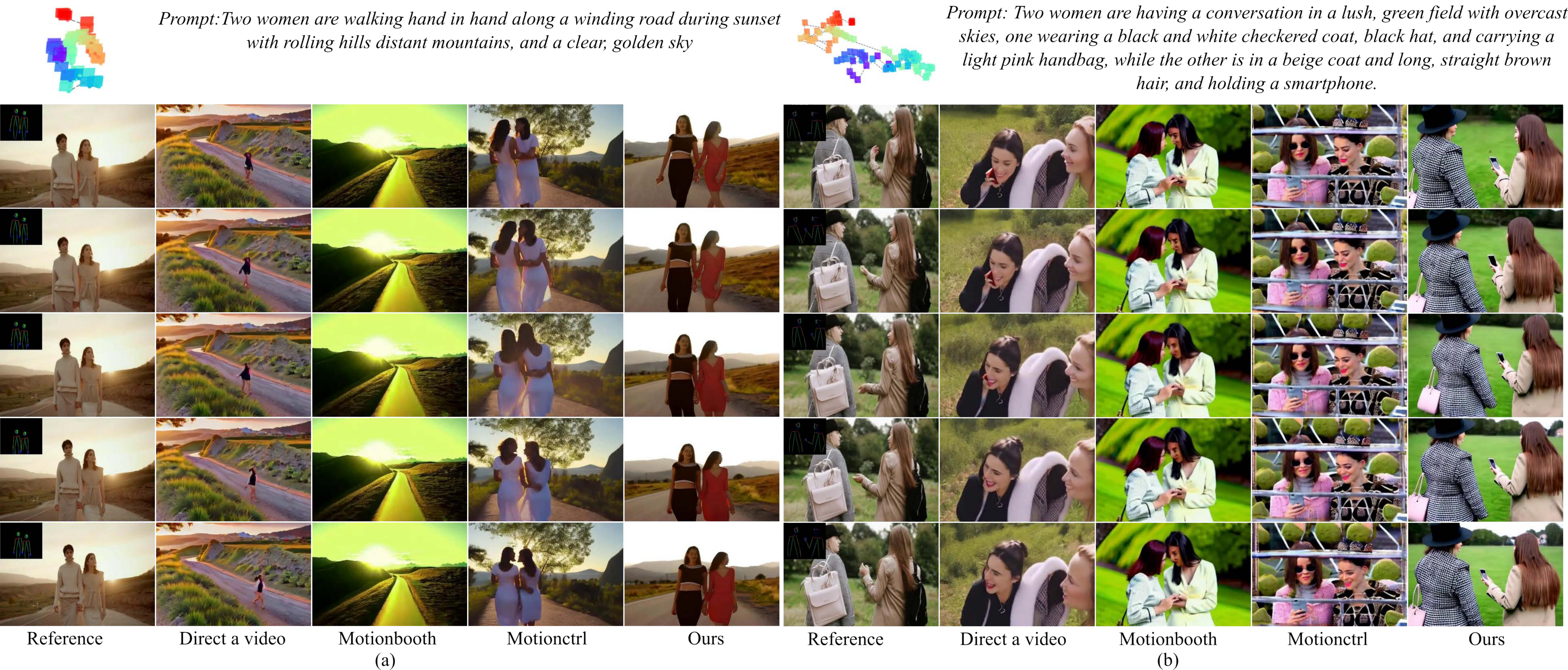}
   \caption{Visualization of joint-control video generation results from Direct-A-Video \cite{DirectAVideo24Yang}, MotionCtrl \cite{MotionCtrl24Wang},   MotionBooth \cite{MotionBooth24Wu} and our TokenMotion-T. Above cases shows that our TokenMotion method succeeds in jointly handling controls of both human motion and camera motion, while being consistently aligned with the input prompts at the same time.}
   \label{fig:case_combine}
\end{figure*}

\subsection{Preliminary}\label{sec3-1: preliminaries}
%Contemporary motion control methodologies predominantly rely on UNet-based diffusion models \cite{SVD23, VideoCrafter1Chen}, extending the established image generation architecture by incorporating temporal attention mechanisms alongside spatial processing. While this frame-sequential approach accommodates basic camera motion and pose guidance, the exclusive reliance on temporal attention presents significant limitations when managing concurrent complex camera movements and human dynamics.
Recent developments in transformer-based diffusion frameworks \cite{CogVideoX24Yang, OpenSoraPlan24Yuan} have markedly advanced the field of video generation, particularly through the synergistic integration of 3D Causal Variational Autoencoders (VAE) with sophisticated spatiotemporal (3D) attention mechanisms. Specifically, the 3D Causal VAE converts a video clip of shape $T \times H \times W \times C$ into sequences of visual tokens $z_{\text{visual}}$ of length $\frac{T}{q} \cdot \frac{H}{p} \cdot \frac{W}{p}$, where $H$, $W$, and $T$ denote the height, width, and length of video, and $p$ and $q$ are spatial and temporal compression ratios, respectively. $z_{\text{visual}}$ is flattened into 1-dimensional vectors and fed into $N$ stacked 3D fully attention blocks, together with positional encoding, text prompt encoding, and timestamps for the denoising process.\\
\indent The video diffusion framework learns a denoiser model $D_{\boldsymbol{\theta}}$ which predict the clean visual tokens from noise inputs $\tilde{z} = z + \sigma \varepsilon, \sigma \in \mathbb{R}_{+}, \boldsymbol{\varepsilon} \sim \mathcal{N}(\mathbf{0}, \boldsymbol{I})$ following the guidance including text prompt $c$ and additional structure control signals $s$ (i.e. camera trajectories and human poses). Consequently, the denoiser model $D_{\boldsymbol{\theta}}$ is parametrized as $D_{\boldsymbol{\theta}}(z_{\text{visual}} ; \sigma)=c_{\text {skip }}(\sigma) z_{\text{visual}}+c_{\text {out }}(\sigma) F_{\boldsymbol{\theta}}\left(c_{\text {in }}(\sigma) z_{\text{visual}} ; c_{\text {noise }}(\sigma)\right)$ and $F_{\boldsymbol{\theta}}$ is network to be trained to optimize the objective : 
\begin{equation}
\mathbb{E}_{\left(z,c, s \right)} \left[\lambda_\sigma\left\|D_{\boldsymbol{\theta}}\left(\tilde{z} ; \sigma, c, s \right)- z\right\|_2^2\right]
\end{equation}

\subsection{Motion Control}\label{sec3-2: MotionControl}
%The implementation of 3D full attention architecture facilitates efficient propagation of control signals across both spatial and temporal dimensions, enabling the learning of global dependencies while maintaining visual fidelity. However, the simultaneous integration of dual control signals—camera motion and human motion—introduces potential control conflicts that must be addressed.To mitigate these challenges, we systematically disentangles camera motion and human motion cues prior to their application to visual tokens. This methodology employs dual specialized feature encoders for the discrete extraction of camera and human motion representations. %While video temporal compression enables the generation of extended sequences, it presents significant challenges in properly tokenizing both camera movements and human motion data in alignment with spatial and temporal compression parameters. %
%Following the extraction of visual tokens via Causal VAE, we implement Causal 3D motion encoders which covert camera and human motion representations into latent features with same spatial and temporal compression ratios with visual tokens. The content-aware motion attention module subsequently processes these camera and human motion tokens to systematically aggregate motion cues. The resulting fused tokens are then utilized to update visual tokens through cross-attention layers. The following subsections provide comprehensive details of each constituent module.

\subsubsection{Camera Motion Representation}\label{sec3-2-1: CameraMotionRepresen}
We follow CameraCtrl~\cite{CameraCtrl24He} to adopt Pl\"ucker embeddings to represent camera pose conditions. For each video clip, its camera condition comprises a sequence of camera parameters, each consisting an intrinsic matrix $\mathbf{K}\in \mathbb{R}^{3\times 3}$, and an extrinsic matrix $\mathbf{P}=[\mathbf{R}; \mathbf{t}]\in \mathbb{R}^{3\times 4}$, where $\mathbf{R}$ and $\mathbf{t}$ denote rotation and translation components, respectively. To effectively anchor the raw values of camera poses within the pixel coordinates, we view camera control signals as camera rays originating from the camera center, represented with Pl\"ucker coordinates given some $\mathbf{K}$ and $\mathbf{P}$. Mathematically, For a random pixel $(u, v)$ in the $f$-th frame, its corresponding ray $\mathbf{p}_{u,v,f}$  in the Pl\"ucker coordinates can be represented as:
\begin{small}
\begin{equation}
    \mathbf{p}_{u,v,f} = \frac{(\mathbf{d}_{u,v, f}, \mathbf{t_f} \times \mathbf{d}_{u,v,f})}{||\mathbf{d}_{u,v,f}\|}, \mathbf{d}_{u,v,f} = \mathbf{R}_f\mathbf{K}_f^{-1} [u, v, 1]^T + \mathbf{t}_f,
\end{equation}
\end{small}
This parametrizes camera trajectories into a $\ddot{\mathbf{P}} \in \mathbb{R}^{6 \times T \times H \times W}$ representation that maintains dimensional consistency with the video's spatiotemporal structure.
For model input latent space transformation, we implement a patchification module that initially compresses spatial dimensions to $\left(\frac{H}{p}\right) \times\left(\frac{W}{p}\right)$ using 2D convolutional layers, followed by 3D Causal convolution layers that compress the temporal dimension to $\left(\frac{L}{q}\right)$, yielding the final camera tokens $z_{\text{camera}}$. The compression ratios $q$ and $p$ are aligned with those of the visual Causal VAE.
%%%%%
%%%%%
\begin{figure*}
  \centering
  \includegraphics[width=0.99\textwidth]{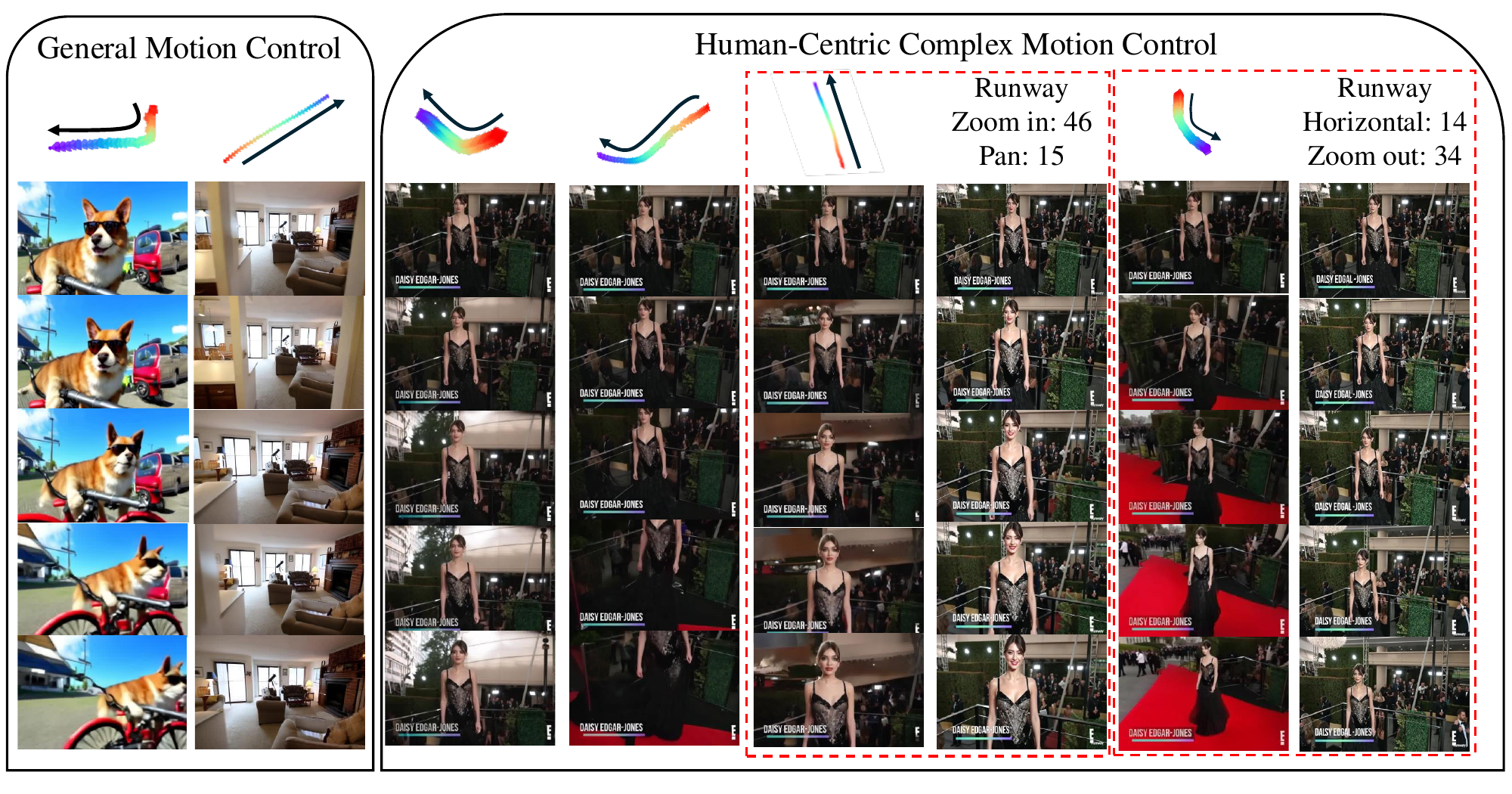}
   \caption{Visualization of camera control over diverse objects, scenarios, and complex camera motion. Our model shows superior visual fidelity and enhanced control flexibility compared to commercial tools (Runway). *Text prompts are omitted to save space.}
   \label{fig:camctrl}
\end{figure*}

\subsubsection{Human Motion Representation}

To condition generated videos on subtle human dynamics, the representation space must have sufficient expressiveness to encode such kinematic subtleties. In light of this, we consider coarse controls, such as bounding boxes and object-level keypoints in previous works~\cite{MotionBooth24Wu, Boximator24Wang}, suboptimal choices, and instead adopt a pose-based representation computed by DWPose~\cite{yang2023effective} for encoding human motion signals. This representation effectively captures subtle changes in facial expressions and postural nuances. Its robustness holds for both single-person and multi-person scenarios, making it an ideal choice for complex human-centric motion control. The same patchification module structure as described in Sec.~\ref{sec3-2-1: CameraMotionRepresen} is then used to encode the human motion conditions.
% Human motion can be represented through multiple modalities ranging from coarse to fine-grained levels, including bounding boxes ~\cite{MotionCtrl24Wang, Boximator24Wang}, and optical flow fields ~\cite{wang2024videocomposer}. In this work, to achieve precise motion control while demonstrating the feasibility of motion decoupling, we adopt a pose-based representation computed using DWPose~\cite{yang2023effective}, which provides a robust foundation for capturing human kinematic dynamics.
% Our empirical evaluation demonstrates this approach's effectiveness and efficiency in extracting human pose cues. 
% The resulting human motion tokens $z_{\text{pose}}$ maintain dimensional consistency with both camera and visual tokens, enabling seamless integration in subsequent processing stages. 
% This architectural design proves crucial for maintaining temporal synchronization and spatial correspondence in the generated videos. %The unified token length across modalities facilitates effective multi-modal fusion operations while preserving the temporal and spatial relationships essential for high-quality video generation. Our experiments validate that this streamlined approach successfully captures and encodes human pose information while maintaining computational efficiency.
%%%%
%%%%

\subsubsection{Decouple-and-Fuse Strategy}
Having encoded camera conditions and human-motion conditions separately, we now focus on enabling the model to learn the interactions between the two motions.  
Intuitively, the model should first project the learned camera conditions globally across the entire latent, then project human-motion conditions only to human-relevant regions, while simultaneously ensuring these localized human poses harmonize with the established camera perspective.

As illustrated in Fig.\ref{fig:framework}, the projection of two motion conditions begins by processing flattened motion tokens through parallel self-attention blocks, enabling the learning of global dependencies that maintain motion consistency across frames.
Then, the localized nature of human-motion conditions compared to the global nature of camera-motion conditions necessitates an implicit decoupling approach that respects their distinct operational domains.
We implement this decoupling with a masking strategy, which aims to enable the model to isolate human-motion effects only to relevant regions. We further relax the learning of this constraint by employing a dynamic mask that combines a learnable component with a hard human-pose prior.  
Specifically, given the camera tokens $z_{\text{camera}}^{n}$ and human pose tokens $z_{\text{pose}}^{n}$ on n-$th$ attention block, we first apply a dimension reduction layer following with a normalization layer to get the learnable parts of the motion masks, $\mathcal{M}_{\text{pose}}^n$ and $\mathcal{M}_{\text{camera}}^n$:

\begin{equation}
\begin{gathered} 
\mathcal{M}_{\text{pose}}^n = \text{LN}(\text{Linear}_{\text{pose}}^{n}(z_{\text{pose}}^{n})) \\ 
\mathcal{M}_{\text{camera}}^n = \text{LN}(\text{Linear}_{\text{camera}}^{n}(z_{\text{camera}}^{n})) \\
\end{gathered}
\end{equation}

We then obtain the human-pose prior $\mathcal{M}_{\text{pose}}^{\text{n, prior}}$ by first resizing the raw human-pose representation to match the latent dimension of the $n$-th block, then dilate it to delineate human-relevant regions, and transform it into a binary mask. $\mathcal{M}_{\text{pose}}^{\text{n, prior}}$ is then added to the learned human-pose mask $\mathcal{M}_{\text{pose}}^n$. We finally compare the token attention mask token-wise with softmax and fuse the tokens accordingly:
\begin{equation}
\begin{gathered} 
z_{\text{fused}}^{n}  = \operatorname{Softmax}\left( [\mathcal{M}_{\text{pose}}^n + \mathcal{M}_{\text{pose}}^{\text{prior}};\mathcal{M}_{\text{camera}}^n] \right) [z_{\text{pose}}^{n}; z_{\text{camera}}^{n}]
\end{gathered}
\end{equation}

We then encode $z_{visual}^n$, the visual tokens on the $n$-th block, with $z_{fused}^n$ via the cross-attention mechanism following a ControlNet~\cite{zhang2023controlnet} block architechture. To effectively enable the model to learn the interaction between two motion conditions, we restrict the cross-attention to only the fused motion condition rather than separate motion conditions, forcing the model to disentangle motions from the fused representation to align with the two conditions for minimizing the distance from the reference videos during training. Specifically, $z_{\text{visual}}^n$ are projected into queries $Q_{\text{visual}}^n$, while the fused motion tokens are transformed into corresponding keys and values $K_{\text{fused}}^n$ and $V_{\text{fused}}^n$ with three learnable linear layers. To maintain optimal balance between motion controllability and video generation fidelity, we incorporate a LoRA (Low-Rank Adaptation) layer prior to updating the visual tokens with the learned motion information. This fusion process can be written as:

% The visual token update process employs cross-attention mechanisms on the fused motion tokens following a ControlNet block architecture. Specifically, at the nth attention block, the original visual tokens $z_{\text{visual}}^n$ are projected into queries $Q_{\text{visual}}^n$, while the fused motion tokens are transformed into corresponding keys and values $K_{\text{fused}}^n$ and $V_{\text{fused}}^n$ with three learnable linear layers. To maintain optimal balance between motion controllability and video generation fidelity, we incorporate a LoRA (Low-Rank Adaptation) layer prior to updating the visual tokens with the learned motion information, following:
\begin{equation}
z_{\text{visual}}^{n}  +=  \text{Lora}(\text{CrossAttn}(Q_{\text{visual}}^n, K_{\text{fused}}^n, V_{\text{fused}}^n))
\end{equation}

This architectural design enables fine-grained control while preserving the fundamental quality characteristics of the generated video sequences.

\begin{table*}[htbp]
    \footnotesize
    \centering
    \setlength{\tabcolsep}{4pt}
    \begin{tabular}{clccccccccccc}
    \toprule
    \toprule
          & \multirow{2}[4]{*}{Method} & \multicolumn{3}{c}{General Metrics} &       & \multicolumn{3}{c}{Camera Metrics} &       & \multicolumn{2}{c}{Human-Motion Metrics} &  \\
\cmidrule{3-5}\cmidrule{7-9}\cmidrule{11-12}          &       & LPIPS$\downarrow$ & FID$\downarrow$   & FVD$\downarrow$   &       & KptsErr$\downarrow$ & RotErr$\downarrow$ & TransErr$\downarrow$ &       & PoseErr$\downarrow$ & DetErr$\downarrow$ &  \\
    \midrule
    \multirow{5}[2]{*}{T2V} & MotionCtrl~\cite{MotionCtrl24Wang} & 0.62 & 116.41 & 1185.83 &       & 6.72  & 0.47  & 8.66  &       & 172.44 & 14.73 &  \\
          & Direct-A-Video~\cite{DirectAVideo24Yang} & 0.67  & 132.08 & 941.05 &       & \textbf{3.85}  & \textbf{0.24}  & \textbf{5.06}  &       & 173.26 & 10.68 &  \\
          & MotionBooth~\cite{MotionBooth24Wu} & 0.69  & 125.62 & 795.59 &       & 5.24  & 0.36  & 6.42  &       & 165.49 & 13.19 &  \\
          & $\text{CogVideoX}^*$-2B~\cite{CogVideoX24Yang}\* & 0.57  & 88.11 & 402.34 &       &   -    &   -    &    -   &      &    -   &    -   &  \\
          & Ours  & \textbf{0.46} & \textbf{82.36} & \textbf{361.03} &       & 4.43  & 0.27  & 5.34  &       & \textbf{45.24} & \textbf{2.50} &  \\
    \midrule
    \multirow{2}[2]{*}{I2V} & ImageConductor~\cite{ImageConductor24Li} & 0.43  & 131.36 & 878.59 &       & 4.32  & 0.44  & 5.44  &       & 164.28 & 9.18  &  \\
          & Ours  & \textbf{0.42} & \textbf{74.65} & \textbf{332.27} &       & \textbf{3.77} & \textbf{0.22} & \textbf{2.98} &       & \textbf{83.82} & \textbf{3.63} &  \\
    \bottomrule
    \bottomrule
    \end{tabular}%
    
    \caption{Quantitative comparisons for joint controlling camera and human motion for both T2V and I2V generation. * denotes the original CogVideoX model, whose camera metrics and human-motion metrics are not calculated because no motion control is performed.}
    \label{tab:main results}
\end{table*}
\section{Experiments}
%
%% TODO: the necessity of having a paragraph like this as different models have different pre-processing proesses and configurations. If we make the statement in all three parts of experiments, it might be a bit repetitive. But I am deleting this as it might be too lengthy.
%
\subsection{Implementation Details} \label{sec4-1: implementation}
We implement two variants of our TokenMotion framework: TokenMotion-T for text-to-video (T2V) and TokenMotion-I for image-to-video (I2V) generation.
Both variants leverage the open-source CogVideoX-2B \cite{CogVideoX24Yang} as the backbone.
Specifically, since CogVideoX-2B only provides the official T2V checkpoint, we implement TokenMotion-I by fine-tuning this checkpoint for 100,000 steps, incorporating input images as an additional conditioning signal.
For both models, we train them on 2 $\times$ 8 A100(80G) GPUs for 5 days using the Adam optimizer \cite{Adam14Kingma}, with a learning rate of 1e-5 and a batch size of 2. During fine-tuning, we only train the two encoders of camera motion and human motion, along with the decouple-and-fuse module, while all the remaining parameters being frozen.  
For training data, we construct a mixed dataset consisting of randomly-sampled 49-frame video clips in a resolution of $256\times 384$, which is a combination of HumanVid \cite{HumanVid24Wang} and RealEstate10K \cite{Realestate18Zhou}. Specifically, HumanVid contains videos with both control signals of human motion and camera motion, while RealEstate10K contains videos with only camera motion signals. To accommodate these differences, we input DWPose-extracted poses for HumanVid samples and blank human-motion cues for RealEstate10K samples during training, respectively. 
The inference is performed with 50-step denoising using DDIM Sampler \cite{DDIM21Song}, with a classifier-free guidance \cite{ho2021classifier} scale of 7.5. More details of preprocessing steps and model configurations are provided in the supplementary.
\subsection{Joint Camera and Human Motion Control}\label{sec3-1: joint}
\paragraph{Evaluation Dataset.} We carefully curated an evaluation set containing 500 real-world videos from HumanVid \cite{HumanVid24Wang} which comprises both camera and human motions. The curation begins with 1000 randomly-sampled videos, from which we removed the 500 videos exhibiting minimal camera motions. This selection complements the focus of HumanVid's real-world set, which prioritizes human motion magnitude with some videos having limited camera dynamics. 
\paragraph{Comparison Methods.} We compare both variants of TokenMotion against state-of-the-art joint-control approaches. For T2V generation, we compare TokenMotion-T with MotionCtrl \cite{MotionCtrl24Wang}, Direct-A-Video \cite{DirectAVideo24Yang} and MotionBooth \cite{MotionBooth24Wu}. For I2V generation, as no existing methods support joint-control, we compare TokenMotion-I with ImageConductor \cite{ImageConductor24Li}, the closest method, which offers only separate control of the two motions. We implement the joint-control in a cascaded manner for ImageConductor. 
% For comparison fairness across methods that use different motion representations. we carefully preprocess all motion signals according to each method's specific requirements.
%
\paragraph{Evaluation Metrics.} We evaluate joint-control of camera motion and human motion from three aspects: (1) General visual quality, for which we use LPIPS \cite{zhang2018lpips}, FID\cite{FID17Heusel} and FVD\cite{FVD19Unt}. (2) Camera movement alignment, for which we follow CameraCtrl \cite{CameraCtrl24He} and CamCO \cite{CamCo24Xu} to extract camera trajectories using COLMAP \cite{schoenberger2016colmap}, which are then leveraged to compute the keypoints error (KptsErr), rotation error (RotError) and transition error (TransErr). (3) Human-motion alignment, for which we follow AnimateAnyone \cite{AnimateDiff24Guo} to use DWPose \cite{yang2023effective} for human pose extraction, and calculate the accuracy for generated motion. This measurement includes the pose-landmark detection failure (DetErr) rate and the Euclidean distances between detectable landmarks of ground-truth and generated poses (PoseErr). 
\paragraph{Quantitative Results.} 
% We conduct quantitative and qualitative comparisons with state-of-the-art methods for joint-control with results provided in Tab.\ref{tab:main results} and Fig.\ref{fig:combine_case}, respectively.
Results shown in Tab.\ref{tab:main results} demonstrate that our approach surpass all baselines in both general visual quality and human-motion alignment in both scenarios of T2V and I2V generation. Specifically, for human-motion control in T2V generation, our TokenMotion-T achieves a pose error of 83.82 and a landmark-detection failure rate of 2.5\%, representing significant improvements of PoseErr and DetErr (164 $\rightarrow$ 83; 9.18\%$\rightarrow$ 3.63\%), which indicates that the model manages to decouple the two motions well. In terms of the general visual quality, our TokenMotion-T also achieves consistent improvements across three metrics, with a significant reduction in FVD from 795.59 to 361.03. This indicates that the unified modeling of human motion and camera motion is effective in achieving the spatiotemporal consistency and the balance between two motions. Although TokenMotion-T achieves worse performance compared to Direct-A-Video, it still outperforms all other baselines. Specifically, Direct-A-Video achieves almost the worst human-motion control across all baselines, indicating that it fails to achieve balance in joint-control. 

For joint-control in I2V generation, our TokenMotion-I consistently outperforms ImageConductor~\cite{ImageConductor24Li} across all metrics, highlighting that controlling human motions and camera motions in a decoupling manner without handling their interactions yields compromised performance, indicating the effectiveness of a decouple-and-fuse strategy proposed in our work.
\paragraph{Qualitative Results.} Fig.\ref{fig:case_combine} further visualizes the superiority of TokenMotion-T with two challenging joint-control scenarios. We adopt TokenMotion-T for qualitative comparisons and  our TokenMotion-T achieves the best performance in following the joint-control signals, while baseline methods fail in different ways: MotionCtrl fails to follow the human motion accurately in the first case while MotionBooth and Direct-A-Video lose more human details in the second case. Some confuse the camera motion with the human motion, such as MotionCtrl~\cite{MotionCtrl24Wang} in Fig.\ref{fig:case_combine} (a) misusing the right-moving camera control as the right-moving of the two people. Some sacrifice the visual quality for motion-control, such as MotionBooth~\cite{MotionBooth24Wu} and Direct-A-Video generating~\cite{DirectAVideo24Yang} unrealistic background scenes in Fig.\ref{fig:case_combine} (b). Meanwhile, our TokenMotion can accurately decouple camera motion from human motion, while maintaining the overall visual quality, the same as Tab.\ref{tab:main results} suggests. 

\begin{figure}
  \centering
  \includegraphics[width=0.9\columnwidth]{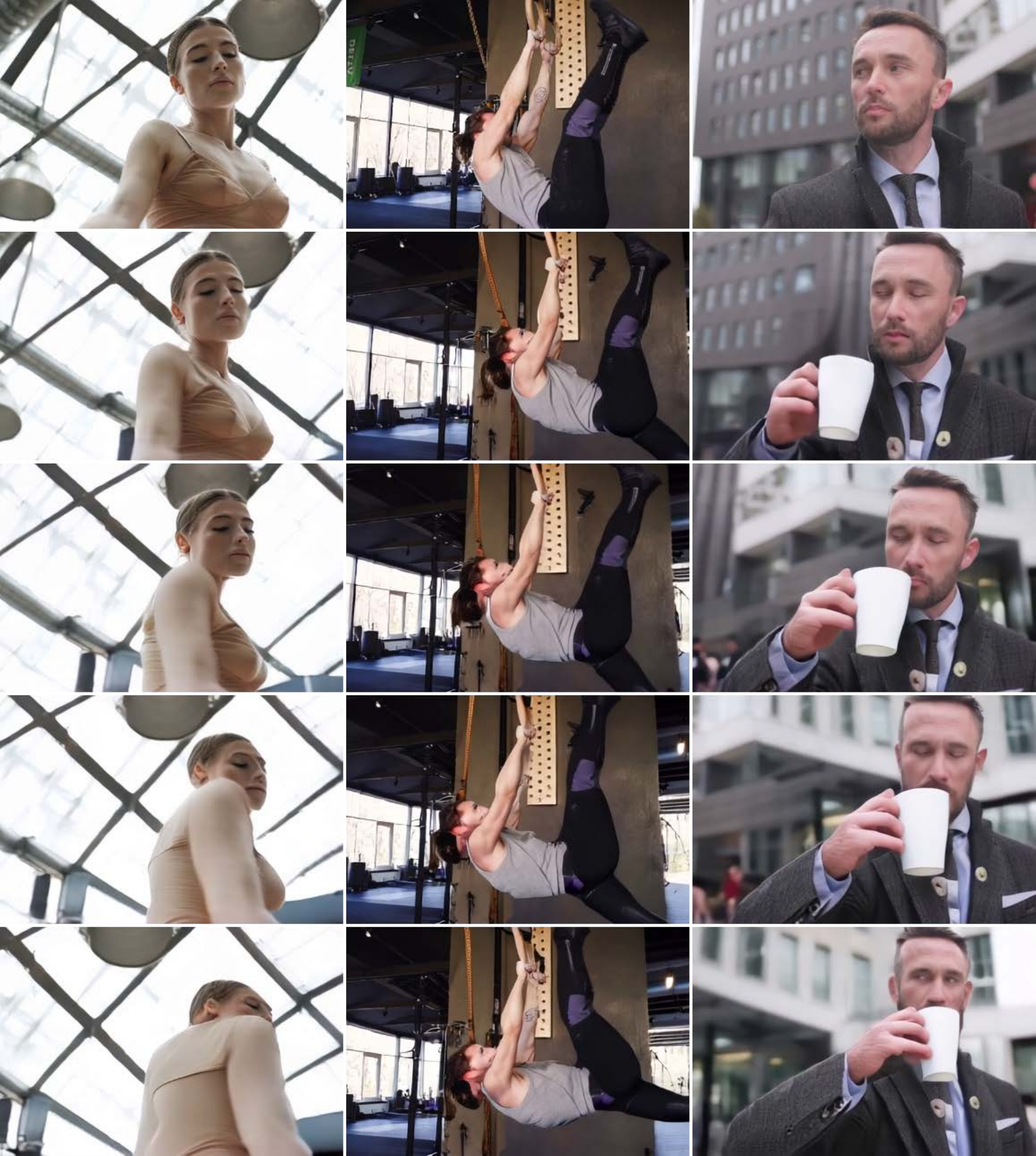}
    \caption{Qualitative results for TokenMotion-I. }
    \label{fig:i2v qualitative results.}
\end{figure}

For TokenMotion-I, Fig.~\ref{fig:i2v qualitative results.} demonstrates that it can perform fine-grained human-motion controls such as body rotation, leg lifting, and eyes blinking simultaneously with complex camera movements such as left-to-right rotation, up-down rotation and zoom-out. Our both TokenMotion variants achieve such advantages beyond these cases, and we refer readers to more generated samples in the supplementary materials for space limitation. 

\begin{table}[htbp]
  \centering
    \begin{tabular}{lccc}
    \toprule
    Methods      & Kpts-err$\downarrow$ & Rot-err$\downarrow$ & Trans-err$\downarrow$ \\
    \midrule
    CameraCtrl~\cite{CameraCtrl24He} & 5.61  & \textbf{0.27} & 6.48 \\
    MotionCtrl-cam~\cite{MotionCtrl24Wang} & 4.46 & 0.33  & 5.83 \\
    Ours  & \textbf{4.36}  & 0.28  & \textbf{5.39} \\
    \bottomrule
    \end{tabular}%
    \caption{Quantitative  evaluation results for camera-only control video generation.}
  \label{tab:camera only results}%
\end{table}%

\subsection{Camera Motion Control} \label{sec3-2: camera-only}
To evaluate the camera motion controllability of our proposed TokenMotion framework, we conducted comprehensive experiments on 1,000 randomly sampled videos from the RealEstate10K~\cite{Realestate18Zhou} test set, following the evaluation protocol established in CameraCtrl~\cite{CameraCtrl24He}. Comparative analyses against state-of-the-art camera-control methods, specifically CameraCtrl~\cite{CameraCtrl24He} and MotionCtrl-camera~\cite{MotionCtrl24Wang}, are presented in Tab. \ref{tab:camera only results}, using camera metrics detailed in Section \ref{sec3-1: joint}. Our method achieves superior performance in Trans-err and Kpts-err, while maintaining competitive performance in Rot-err, demonstrating exceptional camera motion control capabilities. Although TokenMotion is jointly trained on both camera and human motion samples, it exhibits robust motion control across diverse objects and scenarios in both image-to-video (I2V) and text-to-video (T2V) generation tasks, as shown in Fig. \ref{fig:camctrl}. We further validated its effectiveness in complex camera control scenarios, particularly in human-centric cases. Qualitative results in Fig. \ref{fig:camctrl} illustrate that our model successfully synthesizes temporally coherent video sequences that faithfully adhere to physical camera movements. Furthermore, our method demonstrates performance comparable to the commercial Runway Alpha Gen3's camera control system, showcasing both superior visual quality and enhanced control flexibility.

\begin{figure}
  \centering
  \includegraphics[width=1\columnwidth]{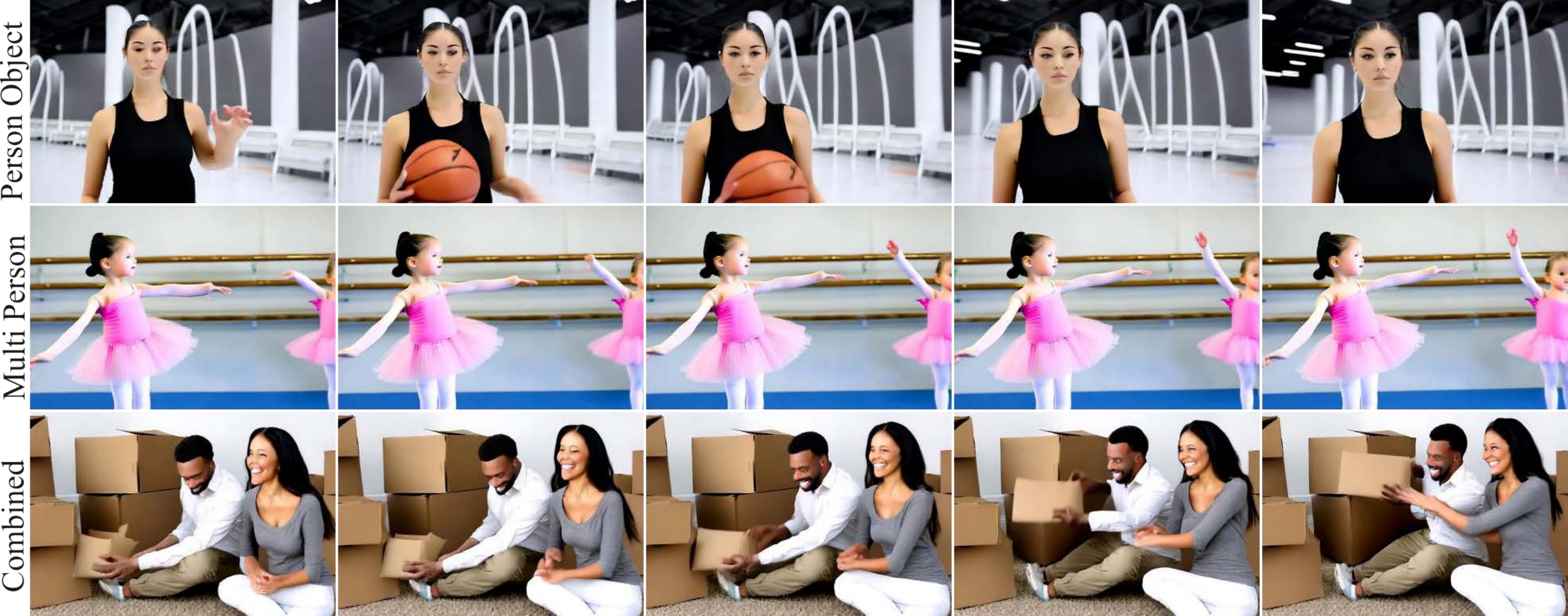}
    \caption{Human-motion control with complex composition.}
    \label{fig:compositional human-motion control.}
\end{figure}

\begin{figure}[htbp]
    \centering
    \begin{subfigure}{\columnwidth}
        \centering
        \begin{tabular}{@{}c@{}c@{}c@{}c@{}c@{}}
            \includegraphics[width=0.2\columnwidth]{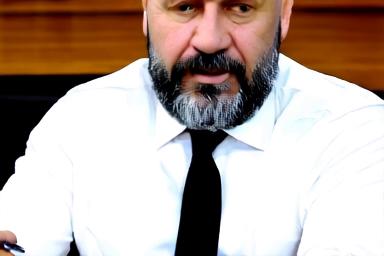} &
            \includegraphics[width=0.2\columnwidth]{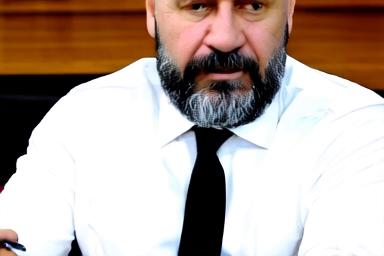} &
            \includegraphics[width=0.2\columnwidth]{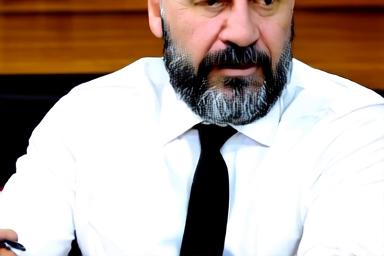} &
            \includegraphics[width=0.2\columnwidth]{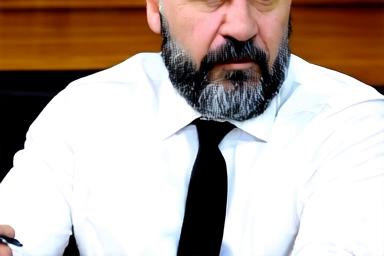} &
            \includegraphics[width=0.2\columnwidth]{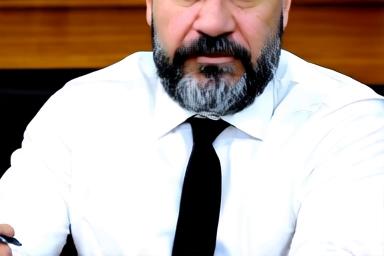} 
        \end{tabular}
        \caption{Human-motion control with subtle facial movements.}
        \label{fig:facial expression}
    \end{subfigure}

    % \vspace{-1mm}  % Adjust space between rows

    \begin{subfigure}{\columnwidth}
        \centering
        \begin{tabular}{@{}c@{}c@{}c@{}c@{}c@{}}
            \includegraphics[width=0.2\columnwidth]{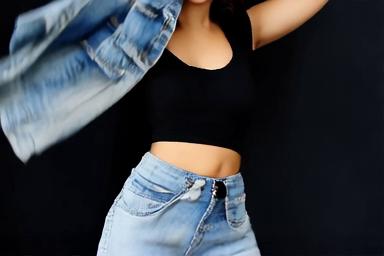} &
            \includegraphics[width=0.2\columnwidth]{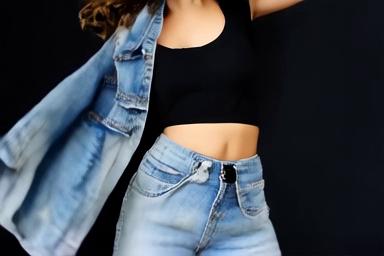} &
            \includegraphics[width=0.2\columnwidth]{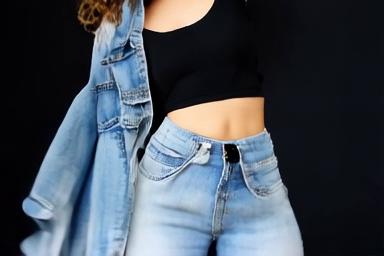} &
            \includegraphics[width=0.2\columnwidth]{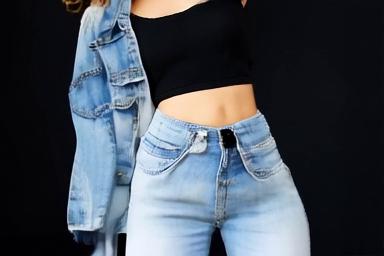} &
            \includegraphics[width=0.2\columnwidth]{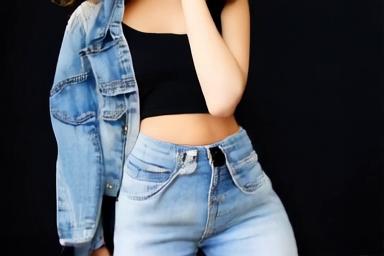} 
        \end{tabular}
        \caption{Human-motion control with large body movements.}
        \label{fig:pose changes}
    \end{subfigure}
    \caption{Control with human motions of different granularity.}
    \label{fig:movements.}
\end{figure}

\subsection{Human Motion Control} \label{sec3-3: human-motion-only}

We show that TokenMotion has strong generalization ability to handle extensive scenarios of human-motion control. As shown in Fig.~\ref{fig:compositional human-motion control.}, TokenMotion manages to perform compositional human motion control, including single-person object interactions, multi-person interactions and multi-person object interactions stably without artifacts. Additionally, TokenMotion is also able to handle human motions of different granularity. While Fig.~\ref{fig: Figure1} has demonstrates that TokenMotion is able to handle full-body movements like trajectories, we additionally present Fig.~\ref{fig:movements.} to show that our model also stably advances in non-full-body movements such as pose changes in Fig.~\ref{fig:pose changes} and even-smaller changes in facial expression, such as lip twitching and head swaying, in Fig.~\ref{fig:facial expression}.

% \paragraph{Evaluation Setups}
% We compare TokenMotion-I with two human image animation methods: AnimateAnyone \cite{AnimateAnyone24Hu} and MagicAnimate \cite{MagicAnimate24Xu}. We follow both their evaluation set and metrics, using the Tiktok dance dataset \cite{TikTok21Jaf}, which contains 340 videos featuring pure human movements under static cameras, and evaluating these methods by SSIM \cite{wang2004ssim}, PSNR \cite{hore2010psnr}, LPIPS \cite{zhang2018lpips}, FID, and FVD. Qualitative results of TokenMotion-T are shown in the supplementary materials.

\subsection{Ablation Study}\label{sec3-4:ablation}

\paragraph{Baseline.} To evaluate the impact of the proposed TokenMotion, we conduct experiments on our base model, CogvideoX, whose results are shown in Tab.~\ref{tab:ablation}. The base model performs worse than our approach in all three metrics, with a significant performance difference in FVD. This indicates the introduced motion signals largely benefit the spatiotemporal consistency of generated videos.

\paragraph{Token Compression.} To demonstrate that the motion signals are effectively encoded, we replace the patchification module in TokenMotion by a ControlNet \cite{zhang2023controlnet} module, which yields motion representations of larger dimensions, the same as hidden states. As shown in Tab.~\ref{tab:ablation}, without patchification, the model produces a worse FVD score, indicating that this model variant has worse motion controllability in generation.

\paragraph{Decouple-and-Fuse.} To demonstrate the effectiveness of our decouple-and-fuse strategy, we conduct experiments on implementing the joint modeling of camera motion and human motion as direct addition. Results in Tab.~\ref{tab:ablation} shows that without the decouple-and-fuse strategy, the model struggles with maintaining both per-frame quality and spatiotemporal consistency in generated videos, as it achieves the worst among all three metrics. This indicates that dedicated-designed modules are necessary for models to handle the interaction between the two motion signals. 

\paragraph{Hybrid-mask Guidance.} To demonstrate the effectiveness of introducing explicit pose guidance in mask, we conduct experiments on a model variant which only leverages a learnable mask for decoupling human motion and camera motion. As shown in Tab.~\ref{tab:ablation}, while a slight performance drop in FVD is shown when incorporating a hybrid mask, scores of LPIPS and FID are much better with our design.

\begin{table}[htbp]
  \centering
    \begin{tabular}{lccc}
    \toprule
          Methods& LPIPS$\downarrow$ & FID$\downarrow$   & FVD$\downarrow$ \\
    \midrule
    Baseline & 0.57  & 88.11 & 402.34 \\
    w/o decouple-and-fuse & 0.66  & 93.89 & 890.39 \\
    w/o token compression & 0.54  & 88.56 & 590.62 \\
    w/o hybrid mask & 0.55  & 89.63 & \textbf{358.76} \\
    Ours  & \textbf{0.46} & \textbf{82.36} & 361.03 \\
    \bottomrule
    \end{tabular}%
    \caption{Ablation studies about Token Compression, Decouple and fusion, and Hybrid-mask.}
  \label{tab:ablation}%
\end{table}%

\subsection{Limitations and Future Works}\label{sec3-5:limitations}
While TokenMotion can precisely generate videos that follow the joint control signals, the visual quality is limited by the base model. 
We identify two representative limitations: First, the model shows difficulties in modeling finger movements. Second, the model struggles with facial detail preservation, where facial features can appear blurred or exhibit geometric distortions. Future work could explore adapting TokenMotion's joint-control strategy to larger-scale backbones. For more concrete discussions on limitations, we refer readers to the supplementary materials.

\section{Conclusions}

This paper presents TokenMotion, the first DiT-based framework for joint motion control, targeting the task of human-centric motion control, which demonstrates superior performance in both text-to-video and image-to-video generation.
Our decouple-and-fuse strategy with the dynamic masking mechanism effectively coheres the control signals of camera motion and human motion.
Comprehensive evaluations demonstrate that our TokenMotion outperforms state-of-the-art approaches in simultaneously controlling camera motions and human motions.
{
    \small
    \bibliographystyle{ieeenat_fullname}
    \bibliography{main}
}


\end{document}